\documentclass{ccjnl}

\title{MobiGPT: A Foundation Model for Mobile Wireless Networks}
\author{ Xiaoqian Qi\inst{1}, Haoye Chai\inst{1,*}, Yong Li\inst{1}\corinfo{haoyechai@mail.tsinghua.edu.cn}}
\receiveddate{Sep. 25, 2019}
\reviseddate{Mar. 24, 2020}
\Editor{Wei Ma}

\address[1]{Department of Electronic Engineering, Beijing National Research Center for Information Science and Technology (BNRist), Tsinghua
University, Beijing, 100084, China}

\usepackage{mathrsfs}

\begin{document}
\maketitle

\begin{abstract}
With the rapid development of mobile communication technologies, future mobile networks will provide vast services and resources for daily commuting, production, daily life, and entertainment. Accurate and efficient forecasting of mobile data (\emph{e.g.,} cell traffic, user behavior, and channel quality) can help operators proactively monitor network state changes, orchestrate wireless resources in a timely manner, and enable rational scheduling of infrastructure and users, thereby improving information supply efficiency and service quality. However, current forecasting paradigms rely on customized designs with tailored models for exclusive data types. These customized approaches undoubtedly increase model complexity and deployment costs under large-scale and heterogeneous access networks involving base stations, users, and wireless channels.
In this paper, we design a foundation model for mobile data forecasting, MobiGPT, adopting a unified model structure capable of forecasting three types of mobile network data: base station traffic, user app usage, and channel quality. Specifically, we propose a soft-prompt learning method to help the model understand the features of different mobile data types, and we introduce a temporal masking mechanism to navigate the model through three forecasting tasks: long-term and long-term predictions, and distribution generation, so as to support various optimization scenarios.
Evaluations on real-world datasets with over 100,000 samples demonstrate that MobiGPT accurately forecasts multiple types of data and tasks. Compared to existing models, MobiGPT shows improvements of 27.37\%, 20.08\%, and 7.27\% in forecasting accuracy, reflecting its strong generalization capability. Furthermore, MobiGPT also exhibits superior zero/few-shots performance in unseen scenarios, with an improvement of over 21.51\%, validating the foundation model's strong transferability.
The code is available at: \url{https://anonymous.4open.science/r/MobiGPT-main/}.
\end{abstract}

\section{Introduction}
\label{sec1}

In recent years, large foundation models have rapidly emerged in the fields of computer vision and natural language processing. These models are reshaping the AI ecosystem by leveraging their robust data processing capabilities, generalization abilities, and zero/few-shot learning. A growing number of vertical industries, such as urban planning, finance, and healthcare, are beginning to develop specialized foundation models tailored to their domains~\cite{zhang2024urban, zhang2024multimodal}. Some studies have even proposed the concept of a world model in pursuit of Artificial General Intelligence (AGI)~\cite{goertzel2014artificial}. Foundational models are proven to be not only capable of comprehensively perceiving real-world data but also possess the ability to forecast future data~\cite{ding2024understandingworldpredictingfuture}.

Mobile networks are inundated with vast amounts of information resources and services, aimed at providing rich service capabilities for heterogeneous user devices~\cite{pernici2006mobile}. With the fast evolution of mobile networks, it will be crucial to understand user behaviors and enhance network performance to better provide real-time and customized information. In this context, Foundational Models (FMs) have the potential to greatly empower future mobile networks. 
Firstly, FMs can accurately predict future dynamics of user and mobile data, allowing operators to proactively optimize and make decisions for network management. 
Second, by leveraging the zero/few-shot capabilities of foundational models (FMs), it is possible to estimate service demand in target areas when constructing mobile networks in regions without historical data, thus improving planning efficiency. Moreover, FMs can model the distribution of user behavior patterns without referring to private data and realize privacy-preserving.
With FMs' ability to simulate multi-dimensional data, analysts can synthesize multi-layered network data in virtual environments to construct Digital Twins (DT) of the mobile network~\cite{nguyen2021digital}, which can generate network data based on different engineering parameters, enabling counterfactual simulations without altering the live networks. Such simulations can be used to assess the effectiveness of various operational strategies, such as PCI planning and user mobility management~\cite{6760605, 10478861}.
However, no such foundation model dedicated to mobile networks has been proposed.

Future mobile networks will integrate cutting-edge technologies across information, operations, and communications, with application scenarios spanning multiple dimensions including air, space, ground, and sea~\cite{xu2023space}. The ubiquitous interconnection of massive devices and the flexible cross-domain service capabilities will generate vast amounts of data. These data naturally support the training of FMs.
Moreover, traditional language large models (LLMs) are challenging to directly apply to mobile networks, where the data, such as service traffic and channel quality, is largely time-series data. Using natural language to describe the complex variability of network data is inherently difficult, which makes it challenging for LLM-based approaches to construct a universal representation of network data.
Therefore, the paper aims to develop a unified time-series foundation model to forecast various types of mobile data. However, constructing such FMs is not straightforward and presents two major challenges.

i). \emph{Generalization of multi-type mobile data}.
Mobile networks are characterized by users, infrastructure, and wireless environments, encompassing various types of data (\emph{e.g.,} user behavior, base station traffic, and channel quality). An ideal FM should be capable of forecasting multiple types of data simultaneously. However, existing research typically focuses on designing customized network models for single data type, such as Base Station (BS) traffic generation~\cite{10836819}, channel quality simulation~\cite{10575638}, or user service prediction~\cite{10.1145/3459637.3482076}. These specialized models inevitably increase deployment costs in real-world networks, and there is a lack of a unified AI model to address the forecasting challenges across diverse data types.
A key research challenge lies in designing a universal backbone that helps the model fully explore the correlation between users, networks, and the wireless environment, which makes the model understand the underlying patterns in multi-type mobile data and improve its generalization ability.

ii). \emph{Adaptability to multiple forecasting tasks}.
In mobile networks, operators need to perform a variety of network optimizations to enhance real-time information delivery, covering aspects such as coverage, throughput, and energy consumption~\cite{9896145, 10242332}. These scenarios typically impose different requirements on data forecasting. For instance, improving user throughput requires predicting short-term network demand to devise resource scheduling strategies~\cite{9923420, 10063388}, which demands the foundation model to handle short-term forecasting tasks. Energy optimization for base stations requires avoiding frequent adjustments to station parameters, calling for long-term energy control~\cite{10390348}, thus requiring the model to handle long-term forecasting tasks. Additionally, when planning networks in new areas without historical data~\cite{10086045}, the model should have the capability to perform generation tasks and fit the potential distribution of network demand.
A critical research challenge lies in designing a training framework for the foundation model that enables it to adapt to a variety of forecasting tasks.

This paper proposes a foundation model for data forecasting in mobile networks, MobiGPT. The model utilizes a combination of transformer and diffusion models as the backbone, aiming to provide a universal approach to forecast various types of network data, including base station/cell traffic, user application traffic, and downlink channels (\emph{i.e.}, RSRP), while supporting diverse long-term/short-term and generation tasks.
Firstly, we design a soft-prompt learning algorithm to help the model understand the data characteristics of cell traffic, user behavior, and wireless channels during training, enabling the generation of multi-type mobile data and addressing the first technical challenge. Afterward, we propose a temporal masking self-supervised learning strategy to guide the model in handling different forecasting tasks, thus addressing the second technical challenge.
Our contributions can be summarized as follows:

$\bullet$ We construct MobiGPT, a foundation model for mobile networks based on the diffusion model, designed to forecast diverse mobile data including base station traffic, user behavior, and channel quality. To the best of our knowledge, MobiGPT represents the first attempt at forecasting multi-dimensional and multi-task mobile network data.

$\bullet$ MobiGPT leverages a customized prompt learning mechanism, where prompts effectively integrate with contextual information to help the model understand the varying characteristics of different types of mobile data. Moreover, we propose a task-oriented temporal masking strategy enabling the model to fully learn the temporal correlations between sequential network data, thus enhancing its ability to adapt to multiple forecasting tasks.

$\bullet$ We collect over 100,000 real-world mobile data samples covering cell traffic, user app usage, and wireless channels to evaluate MobiGPT. The results demonstrate that MobiGPT performs well across a variety of forecasting tasks for different data types with improvements of 27.37\%, 20.08\%, and 7.27\%, highlighting its strong universality and versatility. Moreover, MobiGPT is capable of generating accurate mobile data in unseen scenarios, showcasing its robust zero/few-shot capabilities.

\section{Related work}

\emph{Domain-tailored forecasting models.}
Much literature has been dedicated to modeling and forecasting BS/cell traffic, app usage, and wireless channels. In the early stages of research, researchers typically adopted deterministic modeling approaches, including numerical calculations (\emph{e.g.,} Poisson process~\cite{bonald2006erlang}) and simulation software (NS3, Omnet, \emph{etc.}), to predict changes in mobile network data given environmental information. However, such methods often oversimplify live networks and struggle to predict the complex dynamics of mobile data. With the development of machine learning, researchers have shifted their focus to a data-driven forecasting paradigm with domain knowledge.
For BS/cell traffic, existing models often integrate spatial correlations of urban entities and periodic learning modules to enhance performance. 
Li~\emph{et al}~\cite{10.1145/3586164} combined transformers with GCNs to extract the spatio-temporal dependencies of mobile traffic sequences.
STK-Diff~\cite{10836819} employed a diffusion model to generate traffic using a frequency-attention mechanism. KstDiff constructs spatial relationship graphs by incorporating the distribution of POI, population data, and BS topologies~\cite{10.1145/3589132.3625641}. 
For app usage data, current work primarily focuses on designing solutions integrating user profiles and app usage preferences.
The DoppelGANger and Netshare models were introduced to generate both packet attributes and feature sequences concurrently~\cite{lin2020using, yin2022practical}, to provide reliable packet data for telemetry and anomaly detection. Pan et al.~\cite{wang2020packetcgan} proposed the PacketCGAN model to address the issue of data imbalance in datasets that involve multiple service types from smartphones. CoSEM utilized historical app usage patterns to build latent embeddings, which are then used to predict future app usage~\cite{10.1145/3459637.3482076}.
For channel modeling, existing studies often combine established channel models (\emph{e.g.,} WINNER II, Hata) with environmental information to characterize large/small-scale fading characteristics.
Kanto \emph{et al.}~\cite{10575638} proposed using LSTM to directly learn the temporal features in channel data. The Gradient Boosting Decision Tree (GBDT) model combined engineering data from both the transmitter and receiver, along with the Hata model, to predict RSRP~\cite{9700950}. Tao \emph{et al.}~\cite{Tao2020FeatureEB} manually constructed features from the BS’s transmission engineering data to predict RSRP, including path loss calculations and multi-path signal angle computations.

\emph{Foundation models.} 
These models are particularly effective in multitasking and zero/few-shot learning and have been successfully utilized across a range of specialized fields.
Yang~\emph{et al}~\cite{yang2023fingpt} and  Zhang~\emph{et al}~\cite{Zhang2024TowardsUG} introduced foundation models designed to accomplish a range of specialized tasks, including investment analysis, quantitative research, and urban navigation.
Following this insight, numerous universal models for time-series forecasting foundation models have been developed.
By utilizing existing LLM, TEMPO~\cite{cao2024tempopromtrained}  and Time-LLM~\cite{jin2024timellmtimeseriesforecasting} integrated a prompt mechanism within pre-trained LLMs for long-term forecasting, aligning time-series features with natural language tokens through a reprogramming approach.
In contrast, some techniques do not depend on pre-trained LLM models but instead reconstruct foundational models using transformer-based architectures. 
For instance, LagLLama~\cite{rasul2024lagllamafoundationmodelsprobabilistic}  utilized lag indices to label multi-dimensional periodic features, such as those based on monthly, daily, and hourly intervals.
UniST~\cite{10.1145/3637528.3671662}  utilized a memory network to achieve spatio-temporal predictions specifically in urban environments.
TimeGPT~\cite{garza2023timegpt1} substituted the Feedforward layer in the transformer with CNNs to better capture temporal dependencies. 

Although there are several foundational models in various domains, there is currently no such model tailored for mobile networks, especially for time-series data related to traffic, services, and wireless channels. Most of the existing models are still limited to specified solutions for small-scale datasets. However, as discussed in Chapter 1, in real-world mobile networks, operators often require forecasting solutions for multiple types of data and tasks to enhance network efficiency. To address the challenges of transferring small models across multiple scenarios and reduce the operational costs of model deployment, one promising approach is to design a foundational model for mobile network data forecasting. Therefore, this paper explores potential frameworks for mobile network foundational models and designs masking and prompting mechanisms to ensure forecasting performance.

\section{Problem formulation}

\subsection{Mobile data and forecasting task description}

\begin{figure}[t]
  \centering
  \includegraphics[width=\linewidth]{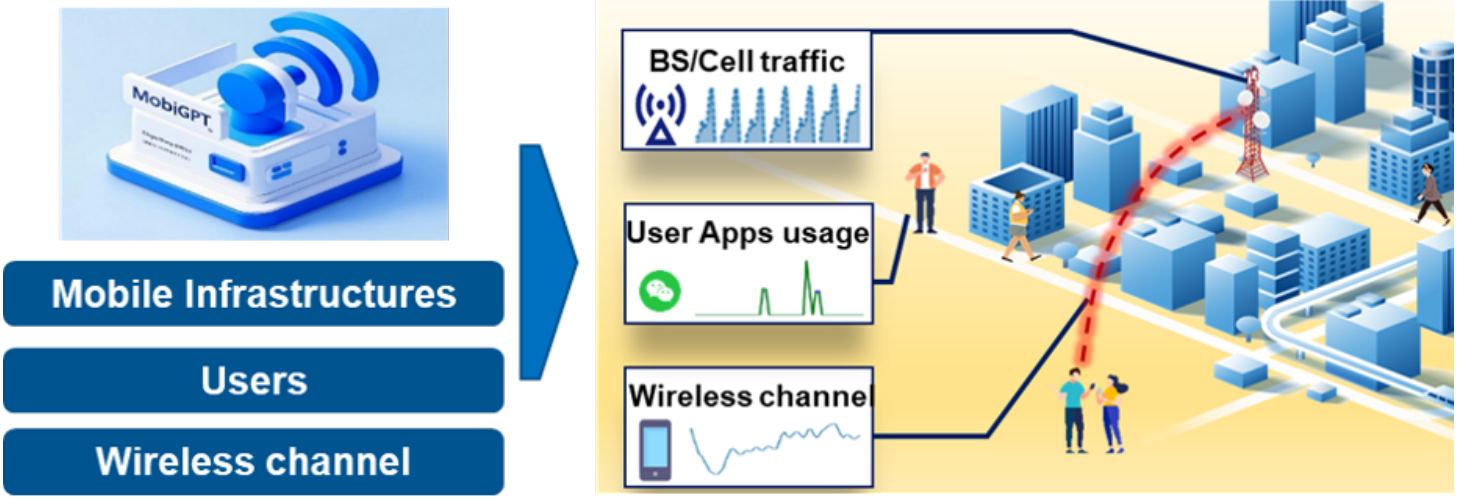}
  \caption{MobiGPT application diagram. MobiGPT is capable of generating various types of data in mobile networks via one unified model.}
  \label{mobigpt_app}
\end{figure}

The mobile network encompasses diverse types of data, including user-side data (\emph{e.g.}, mobility trajectories, app usage sequences, app traffic), wireless channel data (SINR, RSRP, \emph{etc.}), and infrastructure data (\emph{e.g.}, base station traffic, connected users, PRB usage). These data types exhibit distinct temporal variations. This paper aims to design a foundational model that forecasts mobile data from a time-series perspective, as illustrated in Figure~\ref{mobigpt_app}. Specifically, we consider a discrete-time system $\{0,...,l,..., L\}$ with equal time intervals, where, over time, various types of mobile data can be treated as different time series with diverse temporal granularities. For instance, BS traffic is typically reported at 15/30-minute intervals (based on MR reports), user app traffic data is often recorded at a second-level granularity, and channel data is measured at millisecond intervals (based on drive test data). 
As a foundational model for forecasting multiple types of data in mobile access networks, MobiGPT will focus on predicting 3 types of data at the user layer, wireless channel layer, and infrastructure layer, as shown in Figure~\ref{mobigpt_app}.

$\bullet$ BS/cell traffic data in the infrastructure layer: As a typical data type in the infrastructure layer, we denote the traffic at time $l$ as $d^b_l$, which typically exhibits multiple periodicities. Normally, it is strongly correlated with the urban environment including the distribution of Points of Interest (POI), BS topologies, and region functionalities,  which can be denoted as $E^b$. 

$\bullet$ User app traffic in the user layer: It is a representative data type reflecting user behavior and can be denoted as $d^a_l$ at time $l$. 
It often exhibits sparsity in the time domain, generating impulsive traffic data depending on the type of used apps, and is closely linked to individual user profiles and app usage patterns that can be described as $E^a$.

$\bullet$ Downlink RSRP in the wireless channel layer: It serves as a crucial reference indicator for wireless channels, ranging from -40 to -120 dBm and exhibiting high data volatility, and we denote it at time $l$ as $d^c_l$. The data has strong relationships to antenna configuration parameters (power, tilt angle, \emph{etc.}) and the geographical location, which can be described as $E^c$.

\begin{figure*}[t]
  \centering
  \includegraphics[width=\linewidth]{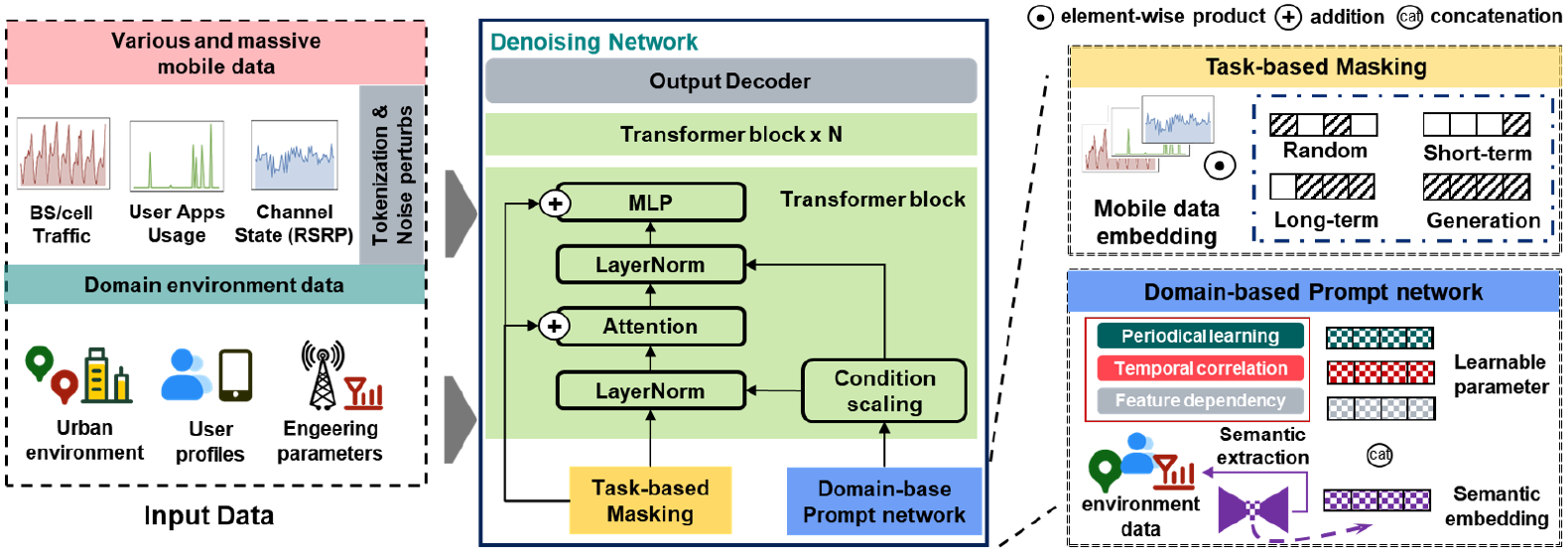}
  \caption{The technical architecture of MobiGPT.}
  \label{foundationmodel}
\end{figure*}

As FMs, MobiGPT not only needs to forecast various types of mobile data but also be capable of handling multiple forecasting tasks, making it applicable to a wide range of network optimization and planning scenarios. Specifically, MobiGPT need to perform the following 3 types of forecasting tasks:

$\bullet$ Short-term prediction. The task relies on historical data $d_{0} \sim d_{l}$ to predict short-term changes in future network data, \emph{i.e.,} $d_{l} \sim d_{L}$, where $l \gg L-l$. By generating short-term predictions, operators can quickly grasp network dynamics and optimize strategies like wireless resource allocation and access control, to enhance user experiences in the short term.

$\bullet$ Long-term prediction. In contrast to the short-term forecasting task, the long-term task focuses on estimating data changes over a longer period,  \emph{i.e.,} $l \ll L-l$, uncovering underlying variation patterns in the data $d_{l} \sim d_{L}$. The task aids operators in evaluating network efficiency from a long-term perspective, facilitating strategic planning for mobile networks, such as cell dormancy and capacity expansion.

$\bullet$ Data generation. Unlike the two above prediction tasks, the generation task does not rely on historical data. Instead, it models the potential distribution of mobile data based on the network environment's factors $E^b, E^a, E^c$, and directly samples forecasting data $d_{0} \sim d_{L}$ from this distribution. The task is especially valuable in new areas with limited or no data, as it enables operators to make decisions on network planning, such as base station placement and digital twin-based inference.

\subsection{Problem formulation}
Based on the preliminaries above, we can define the problem of  foundational model-based mobile data forecasting:

We aim to develop a foundational model $\mathcal{F}$, which can leverage mobile environment features $\mathcal{E}^{(P)}$, and a portion of historical mobile data $\mathcal{Y}^{(L)} = \{ d^b_l, d^a_l, d^c_l\}$ to forecast multi-type and multi-task mobile data $\hat{\mathcal{Y}}^{(L)}$ (BS traffic, app usage, and RSRP)  via one unified structure, \emph{i.e.,} $\mathcal{F}\{\mathcal{E}^{(P)}, Y^{(L)}\} \rightarrow \hat{Y^{(L)}}$, where $\mathcal{E}^{(P)}=\{E^b, E^a, E^c\}$ are the environment features, and $\mathcal{Y}^{(L)}$ and $\hat{\mathcal{Y}}^{(L)}$  correspond to historical reference data and forecast data for different prediction tasks \{short-term, long-term, generation\}.

Building such a foundational model is not straightforward and involves two major technical challenges:
(i). Different forecasting tasks typically require variable-length data inputs. Short-term prediction often needs longer input data, while generation tasks may not require any historical data. \textbf{What} strategies can be employed during the training process to ensure the model can handle such diverse forecasting tasks?
(ii). Different types of mobile data correspond to diverse environmental features. For example, BS traffic is related to POI distribution in the city, while RSRP is influenced by antenna tilt angles, power, and geographical factors. \textbf{How} can we obtain consistent representations of the environmental features of mobile data to ensure the model can forecast various types of mobile data?

\section{Method}

To build the mobile data forecasting foundation model, we propose MobiGPT, as shown in Figure~\ref{foundationmodel}. In general, MobiGPT tokenizes multiple types of mobile data and inputs them along with domain-specific environmental data into the model. Subsequently, MobiGPT utilizes a masking mechanism to identify various forecasting tasks and a prompt network to indicate different types of network data. Afterward, MobiGPT employs diffusion models as the backbone for model training, and we design a conditional scaling mechanism within the transformer block to ensure the model can more accurately capture the correlations between environment conditions and mobile data, thereby enhancing model performance.

\subsection{Input tokenization and noise perturbs}

 Let the input mobile data be $\mathcal{Y}^{(L)} =\{d^b_l, d^a_l, d^c_l \} \in \mathbb{R}^{B\times L\times 1}$, where $B$ is the batch size and $L$ is the temporal length. We utilize temporal convolution to achieve the tokenization process, denoted as $x_0 \in \mathbb{R}^{B\times L\times C_0} = Covn(\mathcal{Y}^{(L)})$, where $C_0$ is the latent embeddings. In this way, the mobile data can be effectively recognized by the transformer structure and input into the task-based masking module.
Subsequently, we apply noise perturbation to the mobile token embeddings in the latent space as:
\begin{equation}
    x_k = \sqrt{\hat{\alpha}_k}x_0 + (1-\hat{\alpha}_k) \epsilon, \ \ \epsilon \sim N(0, I),
    \label{forward}
\end{equation}
where $\alpha_k \in (0,1)$ is a set of scheduled noise weight, $k\in (1,K)$ is the diffusion step, $\hat{\alpha}_k=\prod^k_{k'=1} \alpha_{k'}$, and $\epsilon$ is the Gaussian noise.

To represent the environmental features influencing different types of mobile data, we select 3 distinct domain-specific environmental vectors for the three types of mobile data: For BS traffic, we use urban knowledge graph embeddings~\cite{liu2023urbankg} to capture the spatiotemporal characteristics around the BSs, denoted as $E^b \in \mathbb{R}^{B\times C_1}$; for user app usage, we choose a user profile $E^a \in \mathbb{R}^{B\times L \times C_2}$ reflecting service preferences to characterize user behavior~\cite{10.1145/3597212}; for downlink RSRP, we select BS engineering parameters including antenna tilt, transmission power, and context factors like path distance, land type, \emph{etc.}, to represent large-scale and small-scale fading of signal transmissions, which can be denoted as $E^c \in \mathbb{R}^{B\times L \times C_3}$.

\subsection{Task-based masking module}

To enable the foundational model to understand different forecasting tasks, we employ a masking mechanism that helps the model grasp the correlation residing in the mobile data by the masking-and-reconstructing process. Specifically, we define the following 3 types of masks:

\emph{Short-term and long-term mask}. The schemes mask token $X \in \mathbb{R}^{B \times L \times C}$ on the temporal dimensions over the period $l \sim L$.
Depending on the ratio of $l$ to $L$, the schemes correspond to short/long-term prediction tasks, respectively:
\begin{equation}
    m_{s/l} = \{ 1, \ 0 \sim l \ | \ 0, \ l \sim L\}.
    \label{short_long_pre_mask}
\end{equation}

\emph{Generation mask.}
It completely obscures the temporal period $0 \sim L$ that forces the model to generate the mobile data sequence. Unlike prediction masks that reveal historical data, it enables the model to learn the implicit distribution by only referring to the domain environment features:

\begin{equation}
    m_{g} = \{ 1, \ 0 \sim L \}.
    \label{generation_mask}
\end{equation}

\emph{Random mask.}
It randomly masks the mobile tokens in the temporal dimensions, which aims to capture diverse correlations of the tokens, aiding the model in understanding the complex features of mobile data. Denote $\mathcal{R}(L)$  as randomly choosing items from $L$, the masking strategy yields:
\begin{equation}
    m_{r} = \{1, \ \mathcal{R}(L) \ | \ 0, \ else  \}.
\end{equation}

The proposed masking mechanism not only helps the model capture the intrinsic correlations within the mobile data, but more importantly, it explicitly guides the model in its current forecasting task, enhancing its ability to handle multiple tasks. After applying the masking strategy, the noisy mobile tokens $x_k$ can be denoted as $x_k \leftarrow x_k \odot (1-m) + x_0 \odot m $ and input into the transformer block, where $\odot$ represents element-wise products. 

\subsection{Domain-based prompt network}
The prompt is essential for foundation models with diverse datasets exhibiting significant variations and distributions since it can retrieve the most relevant information on different data types and guide foundation models to be aware of the forecasting data type.
Given the diverse forms and varying semantic information of domain environment data (\emph{e.g.,} $E^b \in \mathbb{R}^{B\times C_1}$ represents the spatiotemporal correlations of urban entities, while $E^c \in \mathbb{R}^{B\times L \times C_3}$ denotes antenna configuration parameters), to ensure the foundation model effectively utilizes these environmental data, we need to achieve unified representation across different types of environment data, while preserving their original semantic information.

\textbf{Data-driven soft prompt}.  We craft trainable parametric vectors $W_{\theta} \in \mathbb{R}^{B\times L\times C_0}$ as the prompt for understanding forecasting data type, where the elements of $W_{\theta}$ are updating during the backpropagation training. 
Based on the input noisy mobile token $x_k$, we design 3 learnable prompts to capture different features within the data. 

\emph{Periodical learning}. It focuses on learning the periodic features of the data, where we extract the most prominent periodic components from the token:
\begin{equation}
    W^p_\theta \leftarrow \mathcal{T}_m[\mathscr{F}(x_k)],
\end{equation}
where $\mathscr{F}$ denote Fourier transformation, and $\mathcal{T}_m$ represents the operation of selecting the maximum $m$ frequency components.

\emph{Temporal correlation}. It is primarily used to learn the temporal correlations between mobile data across time steps. We use a transformer network to perform feature extraction on the mobile tokens $x_k \in \mathbb{R}^{(B\times L\times C_0)}$ along the time dimension $L$:
\begin{equation}
    W^t_\theta \leftarrow \mathcal{H}_L(x_k),
\end{equation}
where $\mathcal{H}_L$ denotes the transformer network along the $L$ dimension.

\emph{Feature dependency}. It learns the correlations between high-dimensional latent features of mobile data at one temporal point. We aim to use this prompt to understand the complex interaction patterns between various environments and mobile traffic in the latent space. Similarly, we use a transformer network to learn the prompt along the feature dimension:
\begin{equation}
    W^f_\theta \leftarrow \mathcal{H}_{C_0}(x_k).
\end{equation}

\textbf{Semantic extraction}. 
To obtain unified representations of each domain environment data, we use Variational Auto Encoder (VAE) networks to extract the semantic information from the environment data. The VAE networks are able to map the data to a specific latent space while preserving the original features. Specifically, before feeding the environment data into the foundation model, we pre-train three types of VAE networks $\Theta^{(P)}$ corresponds to $\mathcal{E}^{(P)}=\{E^b, E^a, E^c\}$. These networks encode the environment data with diverse dimensions into the consistent latent space via MLP layers that extract semantic embeddings $e^{(P)} \in \mathbb{R}^{B \times L \times C_0}$ of environmental features. 
The VAE network fits the mean $\mu_e$ and variance $\sigma_e$ of the data distribution, \emph{i.e.,} $\Theta^{(P)}(\mathcal{E}^{(P)}) =\{ \mu_e, \sigma_e \}$, and samples the latent semantic information via reparameterization operations $\mathcal{R}$ as $e^{(P)} = \mathcal{R}(\mu_e+\sigma_e \epsilon)$. Additionally, the VAE measures the similarity between the latent distribution and normal distributions to ensure the non-regularization issue of the latent distribution. The VAE minimizes the sum of the reconstruction loss and the similarity loss of the input data to ensure the consistency of the environmental semantic features, which can be expressed as
\begin{equation}
    \mathscr{L}_v = M_1(E^{(P)}, \mu_p+\sigma_p \epsilon) + M_2(\mathcal{N}(\mu_p, \sigma_p), \mathcal{N}(0,I)),
\label{loss_vae}
\end{equation}
where $M_1()$ denotes the Mean Squared Error (MSE), and $M_2()$ denotes the Kullback-Leibler (KL) divergence and $\epsilon \in \mathcal{N}(0,I)$.
After the VAEs converge, the networks can map environment data of different shapes to a unified-dimensional representation vector $e^{(P)} \in \mathbb{R}^{B \times L \times C_0}$, which retains the semantic features of the original environment. 
During foundation model training, different types of environment data first pass through the pre-trained VAEs and the latent embedding $e^{(P)}$ are concatenated with data-driven prompt $W_{\theta}$ to obtain the domain-based prompt $p_{\theta}$, \emph{i.e.,} $p_{\theta} = [e^{(P)} \oplus W^p_{\theta}\oplus W^t_{\theta}\oplus W^f_{\theta}]$.

\subsection{Diffusion backbone and training loss}

The denoising backbone of MobiGPT consists of multiple stacked transformer blocks, where noisy mobile token $x_k$ and domain-based prompt $p_\theta$ serve as sequence inputs and control conditions, respectively. To enable effective learning of the correlations between mobile data and environmental features, we resort to an adaptive conditioning scaling strategy. The strategy adaptively scales the layer normalization parameters in the canonical transformer based on environmental features. Compared to directly concatenating or adding the condition information with the raw data, this approach has been shown to offer better computation efficiency and control effective~\cite{peebles2023scalable}. Given the scaling network $\mathcal{S}_\theta$ (where we use MLP networks), the scaling parameters can be denoted as
\begin{equation}
    \alpha_1, \beta_1, \gamma_1, \alpha_2, \beta_2, \gamma_2 =  \mathcal{S}_\theta (p_\theta).
\end{equation}
Afterward, the scaling parameters are input to the layernorm layer to navigate the conditioning process as
\begin{equation}
\begin{aligned}
        x_k' \leftarrow x_k + \alpha_1 \mathcal{A}_\theta(\beta_1 \mathcal{L}(x_k)+\gamma_1) \\
         x_k'' \leftarrow x_k' + \alpha_2 \mathcal{M}_\theta(\beta_2 \mathcal{L}(x_k')+\gamma_2), \\
\end{aligned}
\end{equation}
where $\mathcal{A}_\theta$ and $\mathcal{M}_\theta$ denote the self-attention layer and feedforward layer, respectively, and $\mathcal{L}$ is the layernorm function.
For the training process of MobiGPT, we adopt the self-supervised strategy where the denoising network reconstructs the masked noisy mobile tokens to fit the posterior distribution of the original data, and the ultimate loss function yields
\begin{equation}
    \mathscr{L}_f = \mathbb{E}_{x_0 \sim q(x)}  \bigg \{ ||\epsilon - \epsilon_\theta(x_k,k| p_\theta)||^2 \odot \{m_{s/l}, m_g, m_r\}  \bigg \},
\end{equation}
where $\epsilon_\theta$ is the denoising network. We summarize the pseudocode of MobiGPT in Algorithm~\ref{algo1}.

\begin{algorithm}[ht]
\caption{Training of NobiGPT network.}
\label{algo1}
{\bf Input:} 
Various mobile data $Y =\{d^b_l, d^a_l, d^c_l\} \in \mathbb{R}^{B \times L \in C_0}$; Diffusion step $K$, Noise schedule $\{\hat{\alpha}_k\}_{1:K}$; Domain environment data: $E^b \in \mathbb{R}^{B \times C_1}$, $E^a \in \mathbb{R}^{B \times L\times C_2}$, $E^c \in \mathbb{R}^{B \times L \times C_3}$\\
{\bf Output:} 
Predicted noise $\epsilon_\theta$
\begin{algorithmic}[1]
\WHILE{Semantic extraction not converged} 
\STATE
Obtain environment latent embeddings: \\
$\mu_e^b, \sigma_e^b =\Theta^b(E^b)$; \\
$\mu_e^a, \sigma_e^a =\Theta^a(E^a)$; \\
$\mu_e^c, \sigma_e^c =\Theta^c(E^c)$;  \\
Take the gradient step  $min \mathscr{L}_v$ in Eq.~(\ref{loss_vae}).\\
\ENDWHILE 
\WHILE{Denoising network not converged} 
\STATE
Randomly choose mask $m = \{m_{s/l}, m_g, m_r\}$; \\
Calculate position embeddings $pe_\theta(L)$; \\
Calculate diffusion step embeddings $pe_\theta(k)$; \\
Tokenization and noise perturbation $x_k$ in Eq.~(\ref{forward}); \\
Calculate masked mobile token: \\
$x_k \leftarrow x_k \odot (1-m) + x_0 \odot m $; \\
Acquire input mobile token: \\
$x_k = x_k +pe_\theta(L) + pe_\theta(k); $ \\
Acquire domain prompts $p_{\theta} = [e^{(P)} \oplus W_{\theta}]$; \\
Take the gradient step: \\
$\nabla_{\theta}[||\epsilon - \epsilon_\theta(x_k,k| p_\theta)||^2 \odot m]$
\ENDWHILE 
\end{algorithmic}
\end{algorithm}

\section{Evaluation}
We conduct evaluations on MobiGPT using over 100,000 real-world datasets, including cell traffic, app traffic, and channel RSRP data, with more than 17 baselines across domain-tailored and universal baselines. Our evaluation addresses the following three key questions:

\textbf{RQ1}. How effectively is MobiGPT forecasting across various data types and forecasting tasks?

\textbf{RQ2.} What is the role of the prompt network, and how does MobiGPT perform in terms of transferability?

\subsection{Evaluation setup}
\emph{Datasets}. \textbf{Cell traffic dataset.} The dataset contains hourly traffic data from over 10,000 BSs across four cities in China, \emph{i.e.,} Nanchang, Beijing, Shanghai, and Nanjing~\cite{10836819}, spanning one month with a total length of 672 time points.
\textbf{App traffic dataset.} Collected by operators in Shanghai, the dataset includes the service sequences and corresponding traffic data for 872 users over one week. The data granularity is at the second level, with a total length exceeding 1,000,000 entries~\cite{10.1145/3597212}.
\textbf{RSRP Dataset}. This dataset consists of over 5,000,000 RSRP measured data points collected from 472 real-world BSs~\cite{4ba2-tg21-20}.
We process these three types of raw data into time-series data with a length of 64 for each sequence, constructing a mixed-type dataset comprising over 100,000 samples in total.

\emph{Baselines}. We select 12 domain-tailored baselines. Moreover, we choose 5 universal time-series forecasting models.

BS traffic forecasting. \textbf{KG-RGCN}~\cite{10.1145/3589132.3625569} utilizes autoregressive approach conditioning on multi-attribute land graphs for long/short-term forecasting tasks. \textbf{5GTGAN}~\cite{10449458} uses a GAN architecture with a Bi-LSTM to model the traffic data distribution, allowing for the generation task. \textbf{KstDiff}~\cite{10.1145/3589132.3625641} integrates urban knowledge graphs and pre-trains an MLP-based traffic average predictor to achieve long/short-term prediction and generation tasks. \textbf{STK-Diff}~\cite{10836819} employs urban knowledge graphs and a frequency-domain attention mechanism for both long/short-term prediction and generation tasks.

App usage forecasting. \textbf{CoSEM}~\cite{10.1145/3459637.3482076} associates user history app usage sequences for preference analysis, enabling both long/short-term prediction tasks. \textbf{NetShare}~\cite{yin2022practical} incorporates service usage sequences as metadata and network packet traces as measurements for generation task. \textbf{PacketCGAN}~\cite{wang2020packetcgan} is a conditional generative network based on GANs with one-hot service encoding, capable of performing both long/short-term prediction and generation tasks. \textbf{MRel-HGAN}~\cite{10.1145/3597212} constructs high-dimensional user app features by considering user geography, social connections, and app usage time, enabling both long/short-term prediction and generation tasks.

\begin{table*}[ht]
\centering
\caption{Performance of short-term prediction task. Bold numbers denote the best results and $\underline{underline}$ numbers denote the second-best results.}
\label{shortterm}
\renewcommand{\arraystretch}{1.2}
{
\resizebox{0.95\textwidth}{!}{
\begin{tabular}{c|ccccccccccc}
\hline
\textbf{Model} & \multicolumn{3}{c|}{\textbf{BS traffic}}                              & \multicolumn{1}{c|}{\textbf{Model}}  & \multicolumn{3}{c|}{\textbf{App traffic}}                              & \multicolumn{1}{c|}{\textbf{ Model}} & \multicolumn{3}{c}{\textbf{RSRP}}          \\ \hline
             \textbf{Metric}          & \textbf{JSD} & \textbf{MAE} & \multicolumn{1}{c|}{\textbf{NRMSE}} & \multicolumn{1}{c|}{\textbf{}}           & \textbf{JSD} & \textbf{MAE} & \multicolumn{1}{c|}{\textbf{NRMSE}} & \multicolumn{1}{c|}{\textbf{}}              & \textbf{JSD} & \textbf{MAE} & \textbf{NRMSE} \\ \hline
\textbf{KG-RGCN}       & 0.0601      & 0.1288     & \multicolumn{1}{c|}{0.2441}         & \multicolumn{1}{c|}{\textbf{CoSEM}}      & 0.3349      & 0.1709      & \multicolumn{1}{c|}{0.3009}         & \multicolumn{1}{c|}{\textbf{BPNN}}          & 0.00108      & \underline{0.1584}       & \underline{0.1196}       \\
\textbf{KstDiff}       & 0.0502      & 0.1303      & \multicolumn{1}{c|}{0.2512}         & \multicolumn{1}{c|}{\textbf{PacketCGAN}} & 0.2488       & 0.1655      & \multicolumn{1}{c|}{0.2774}         & \multicolumn{1}{c|}{\textbf{ChannelLSTM}}          & 0.00200     & 0.1687     & 0.1638        \\
\textbf{STK-Diff}      & \underline{0.0392}       & \underline{0.0977}      & \multicolumn{1}{c|}{\underline{0.2303}}         & \multicolumn{1}{c|}{\textbf{MRel-HGAN}}  & \underline{0.2099}       & 0.1328      & \multicolumn{1}{c|}{0.2677}         & \multicolumn{1}{c|}{\textbf{GBDT}}   & \underline{0.00095}       & 0.1658       & 0.1327        \\ \hline
\textbf{TEMPO}       & 0.0477       & 0.1674       & 0.2690                              &     ---                                     & 0.5106       & 0.2140       & 0.2005                              &       ---                                      & 0.00314       & 0.1761       & 0.4496         \\
\textbf{Lagllama}      & 0.0568      & 0.0988      & 0.2327                            &     ---                                     & 0.5661     & \underline{0.1164}      & \underline{0.1941}                              &      ---                                       & 0.00188       & 0.1679      & 0.2161        \\
\textbf{TimeGPT}       & 0.0544      & 0.1009      & 0.2382                             &    ---                                      & 0.4582       & 0.1891      & 0.2012                             &         ---                                    & 0.00241     & 0.1632      & 0.1327        \\ \hline
\textbf{MobiGPT (our)} & 0.0275      & 0.0838      & 0.2126        &         ---          & 0.1483      & 0.0950      & 0.1838          &       ---              & 0.00089      & 0.1309      & 0.1120        \\ \hline
\textbf{Improvement (\%)}   & 29.85     & 14.23       &  7.68     & ---                   & 29.34      & 18.38       & 5.30      & ---                     & 6.31     & 17.36      & 6.35         \\ \hline
\end{tabular}
}
}
\end{table*}

Channel forecasting. \textbf{GBDT}~\cite{9700950} uses decision trees with transmission engineering parameters and theoretical path loss inputs, which achieves long/short-term prediction tasks. \textbf{WirelessVAE} leverages VAE to encode users, BSs, and network key performance indicators to represent environmental information, which achieves the RSRP generation task as maximum likelihood modeling. \textbf{BPNN}~\cite{Tao2020FeatureEB} reconstructs BS engineering parameters and incorporates weather and terrain factors to obtain multiple features that characterize wireless signal transmission, and \textbf{ChannelLSTM}~\cite{10575638}  treats the channel as a time-series sequence and uses LSTM to learn channel variation features, the two baselines enable both long/short-term prediction and generation tasks.

Foundation/universal model for time-series data. \textbf{TEMPO}~\cite{cao2024tempopromtrained} utilizes temporal prompts with trend and seasonal features for pre-trained models (GPT-2) to predict time series.. \textbf{Lagllama}~\cite{rasul2024lagllamafoundationmodelsprobabilistic} uses a set of lag indices to capture different periodic correlations in the time series, enabling long/short-term prediction tasks.
\textbf{TimeGPT}~\cite{garza2023timegpt1} replaces the feedforward layer in the transformer with a CNN network and is trained on large-scale temporal data, enabling long/short-term prediction tasks.
\textbf{MTS-CGAN}~\cite{madane2022transformerbasedconditionalgenerativeadversarial} adopts a WGAN structure to learn the distribution of multi-type time-series data for the generation task.
\textbf{CSDI}~\cite{10.5555/3540261.3542161} is a conditional diffusion model that employs a masking method for the generation task.

\emph{Metrics.}  Given the generated data $Y$ and the real data $\hat{Y}$, we choose 3 metrics to evaluate our MobiGPT.

\textbf{JSD} (Jensen-Shannon Divergence) is used to quantify the similarity between two probability distributions, which calculates the mean of the Kullback-Leibler (KL) divergence between two distributions:

\begin{footnotesize}
\begin{equation}
    JSD =\sqrt{[KL(Pr(\hat{Y}),Pr(Y))+KL(Pr(Y),Pr(\hat{Y})]/2}.
\end{equation}
\end{footnotesize}

\textbf{MAE} (Mean Absolute Error) evaluates the similarity of generated values $Y$ and real values $\hat{Y}$:
\begin{footnotesize}
\begin{equation}
    MAE = Avg(|Y-\hat{Y}|).
\end{equation}
\end{footnotesize}

\textbf{NRMSE} (Normalized Root Mean Squared Error) measures the difference between predicted values and actual values with normalization:

\begin{footnotesize}
\begin{equation}
    NRMSE = \sqrt{Avg(|Y-\hat{Y}|^2)} / (max(\hat{Y})-min(\hat{Y})).
\end{equation}
\end{footnotesize}

\subsection{Forecasting performance (RQ1)}

\begin{table*}[ht]
\centering
\caption{Performance of long-term prediction task. Bold numbers denote the best results and $\underline{underline}$ numbers denote the second-best results.}
\label{longterm}
\renewcommand{\arraystretch}{1.2}
{
\resizebox{0.95\textwidth}{!}{
\begin{tabular}{c|ccccccccccc}
\hline
\textbf{Model} & \multicolumn{3}{c|}{\textbf{BS traffic}}                              & \multicolumn{1}{c|}{\textbf{Model}}  & \multicolumn{3}{c|}{\textbf{App traffic}}                              & \multicolumn{1}{c|}{\textbf{ Model}} & \multicolumn{3}{c}{\textbf{RSRP}}          \\ \hline
             \textbf{Metric}          & \textbf{JSD} & \textbf{MAE} & \multicolumn{1}{c|}{\textbf{NRMSE}} & \multicolumn{1}{c|}{\textbf{}}           & \textbf{JSD} & \textbf{MAE} & \multicolumn{1}{c|}{\textbf{NRMSE}} & \multicolumn{1}{c|}{\textbf{}}              & \textbf{JSD} & \textbf{MAE} & \textbf{NRMSE} \\ \hline
\textbf{KG-RGCN}       & 0.0590      & 0.1317     & \multicolumn{1}{c|}{0.2413}         & \multicolumn{1}{c|}{\textbf{CoSEM}}      & 0.3114      & 0.0401      & \multicolumn{1}{c|}{0.1996}         & \multicolumn{1}{c|}{\textbf{BPNN}}          & 0.00122     & \underline{0.1340}      & 0.1297        \\
\textbf{KstDiff}       & \underline{0.0421}       & 0.1211       & \multicolumn{1}{c|}{0.2013}         & \multicolumn{1}{c|}{\textbf{PacketCGAN}} & 0.2233     & 0.0311       & \multicolumn{1}{c|}{0.1872}         & \multicolumn{1}{c|}{\textbf{ChannelLSTM}}          & 0.00176      & 0.1546      & 0.1434         \\
\textbf{STK-Diff}      & 0.0402       & 0.1112       & \multicolumn{1}{c|}{0.1999}         & \multicolumn{1}{c|}{\textbf{MRel-HGAN}}  & \underline{0.1812}      & \underline{0.0298}     & \multicolumn{1}{c|}{\underline{0.1795}}         & \multicolumn{1}{c|}{\textbf{GBDT}}   & \underline{0.00107}    & 0.1407     & \underline{0.1296}       \\ \hline
\textbf{TEMPO}       & 0.0720       & 0.1901      & 0.4435                              &     ---                                     & 0.5625       & 0.1239       & 0.2138                              &       ---                                      & 0.00166      & 0.1434      & 0.2520         \\
\textbf{Lagllama}      & 0.0524      & \underline{0.0922}       & \underline{0.1955}                              &     ---                                     & 0.5996      & 0.0578       & 0.2396                             &      ---                                       & 0.00507      & 0.1503       & 0.2129        \\
\textbf{TimeGPT}       & 0.0670      & 0.1231       & 0.2042                            &    ---                                      & 0.6637      & 0.0372     & 0.3532                              &         ---                                    & 0.00223       & 0.1752     & 0.1734        \\ \hline
\textbf{MobiGPT (our)} & 0.0281      & 0.0906     & 0.1739       &         ---          & 0.1273     & 0.0225     & 0.1397         &       ---              & 0.00105     & 0.1004      & 0.1200       \\ \hline
\textbf{Improvement (\%)}   & 30.09      & 1.73      &  11.04     & ---                   & 29.74       & 24.49      & 22.17     & ---                     & 1.87      & 25.07     & 7.41      \\ \hline
\end{tabular}
}
}
\end{table*}

\begin{table*}[ht]
\centering
\caption{Performance of generation task. Bold numbers denote the best results and $\underline{underline}$ numbers denote the second-best results.}
\label{generation}
\renewcommand{\arraystretch}{1.2}
{
\resizebox{0.95\textwidth}{!}{
\begin{tabular}{c|ccccccccccc}
\hline
\textbf{Model} & \multicolumn{3}{c|}{\textbf{BS traffic}}                              & \multicolumn{1}{c|}{\textbf{Model}}  & \multicolumn{3}{c|}{\textbf{App traffic}}                              & \multicolumn{1}{c|}{\textbf{ Model}} & \multicolumn{3}{c}{\textbf{RSRP}}          \\ \hline
             \textbf{Metric}          & \textbf{JSD} & \textbf{MAE} & \multicolumn{1}{c|}{\textbf{NRMSE}} & \multicolumn{1}{c|}{\textbf{}}           & \textbf{JSD} & \textbf{MAE} & \multicolumn{1}{c|}{\textbf{NRMSE}} & \multicolumn{1}{c|}{\textbf{}}              & \textbf{JSD} & \textbf{MAE} & \textbf{NRMSE} \\ \hline
\textbf{5GTGAN}        & 0.0877      & 0.2049      & \multicolumn{1}{c|}{0.3022}         & \multicolumn{1}{c|}{\textbf{NetShare}}   & 0.4022       & 0.1112      & \multicolumn{1}{c|}{\underline{0.1433}}         & \multicolumn{1}{c|}{\textbf{BPNN}}           & \underline{0.00113}       & \underline{0.1078}      & \underline{0.1094}       \\
\textbf{KstDiff}       & \underline{0.0532}    & \underline{0.1265}      & \multicolumn{1}{c|}{\underline{0.2101}}         & \multicolumn{1}{c|}{\textbf{PacketCGAN}} & 0.3999      & 0.1010       & \multicolumn{1}{c|}{0.1501}         & \multicolumn{1}{c|}{\textbf{ChannelLSTM}}          & 0.00131      & 0.1481      & 0.1142         \\
\textbf{STK-Diff}      & 0.0603       & 0.1300      & \multicolumn{1}{c|}{0.2263}         & \multicolumn{1}{c|}{\textbf{MRel-HGAN}}  & \underline{0.3411}       & \underline{0.0996}      & \multicolumn{1}{c|}{0.1533}         & \multicolumn{1}{c|}{\textbf{WirelessVAE}}   & 0.00129      & 0.1267       & 0.1181        \\ \hline
\textbf{MTS-CGAN}      & 0.0983      & 0.1446      & 0.5223                             &          ---                                & 0.6723      & 0.1082       & 0.1474                              &         ---                                    & 0.00200       & 0.1374       & 0.1972        \\
\textbf{CSDI}          & 0.0671       & 0.1381       & 0.2229                             &       ---                                   & 0.5811       & 0.1385      & 0.1766                             &        ---                                     & 0.00152     & 0.1355      & 0.2224         \\ \hline
\textbf{MobiGPT (our)} & 0.0414       & 0.0954      & 0.2003                            &       ---                                   & 0.2847     & 0.0817      & 0.1373                             &         ---                                    & 0.00108      & 0.0998     & 0.1006        \\ \hline
\textbf{Improvement (\%)}   & 22.18      & 24.58       & 4.66                              &     ---               & 16.53       & 17.97      & 4.18                            &     ---                                        & 4.42      & 7.42      & 8.04         \\ \hline
\end{tabular}
}
}
\end{table*}

For the three types of data: BS traffic, app usage, and RSRP, we test the performance of the MobiGPT on short-term prediction task, long-term prediction task, and generation task, as shown in Tables~\ref{shortterm}, \ref{longterm}, and \ref{generation}. Overall, MobiGPT achieves the best forecasting accuracy across various tasks. For traffic forecasting, MobiGPT improves the JSD metric by 29.85\%, 30.09\%, and 22.18\% in terms of the short-term, long-term, and generation tasks, respectively, with an average improvement of 27.37\%. For app traffic, the MAE metric improves by 18.38\%, 24.49\%, and 17.97\%, with an average improvement of 20.08\%. For the wireless channel (RSRP), MobiGPT achieves improvements in the NRMSE metric of 6.35\%, 7.41\%, and 8.04\% across the three tasks, with an average improvement of 7.27\%.
It is evident that the model shows greater improvement in user traffic (whether BS or app traffic) than in wireless channel quality (RSRP). We believe this is because mobile traffic, whether from base stations or apps, reflects human network activity patterns. For example, BS traffic increases during midday and evening and decreases at night, while app traffic is related to the app type and usage sequence. These patterns reflect the spatiotemporal regularities of human activity, which our large model can effectively learn within the mobile and environment data, thus enhancing forecasting performance. On the other hand, wireless channel quality is influenced by various factors such as geography, weather, and urban infrastructures, and user mobility also impacts signal transmission, which contributes to both large-scale and small-scale fading characteristics.  AI models are still limited in understanding and learning the variation of electromagnetic signals in complex environments.

Comparing various time-series foundational models, we observe that while these models are capable of handling multiple forecasting tasks, their overall prediction accuracy is inferior to that of tailored models. The LLM-based foundational model (TEMPO) can forecast multiple data types but performs poorly in terms of forecasting accuracy. For instance, in Tables~\ref{shortterm} and \ref{longterm}, TEMPO achieves MAE values of 0.1674 and 0.1901 for BS traffic forecasting, which are significantly higher than those of other baselines. This is because natural language is not well-suited to accurately describe the diverse features of time-series data, making it more reasonable to build foundational models that focus primarily on time-series data.
Additionally, time-series-based foundational models (\emph{e.g.,} Lagllama and TimeGPT) typically use autoregressive methods for prediction, primarily learning the temporal patterns in historical data, but they lack the ability to explore the dependencies between temporal patterns and environmental factors. This also explains why time-series models perform better in BS traffic tasks with prominent temporal features (for example, in Table~\ref{longterm}, Lagllama achieves a MAE of 0.0992 for BS traffic, which has the second-best performance), but shows poor performance in forecasting app traffic and RSRP data (in Table~\ref{shortterm}, Lagllama's JSD metric for app traffic is only 0.5661, while MRel-HGAN achieves 0.2099).
Unlike existing foundational models, MobiGPT incorporates a semantic extractor for environmental features within the framework. The component extracts semantic embeddings for the environmental features corresponding to different types of mobile data, helping the model capture environmental dependencies. As a result, MobiGPT can effectively improve the forecasting accuracy of domain mobile data while maintaining generalizability.

\begin{figure}[t]
  \centering
  \includegraphics[width=\linewidth]{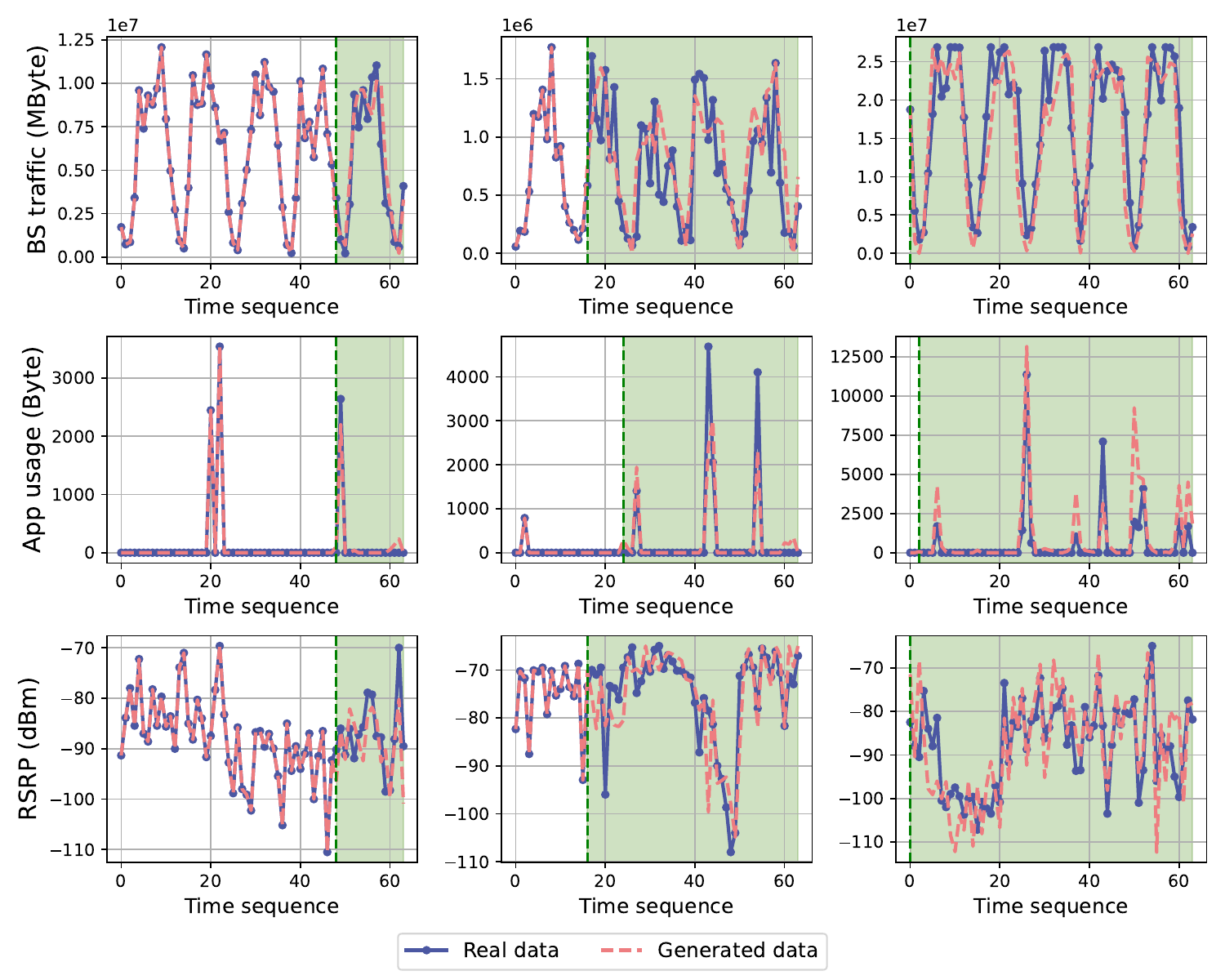}
  \caption{Visualization results of MobiGPT for different data and tasks. From left to right are short-term prediction, long-term prediction, and generation tasks, respectively.}
  \label{visual}
\end{figure}

To visually demonstrate MobiGPT's performance in forecasting multiple types of data, we present a portion of the predicted and real data in Figure~\ref{visual}. From top to bottom, the figure represents BS traffic, app traffic, and RSRP; from left to right, it shows short-term prediction, long-term prediction, and the generation task. It is evident that MobiGPT effectively learns the temporal patterns of various data types, proving the model's effectiveness and universality.

\subsection{Zero/few shot, ablation, and scaling performances (RQ2)}

\begin{figure}[t]
    \centering
    \subfloat[Generation task.]{\includegraphics[width=0.45\linewidth]{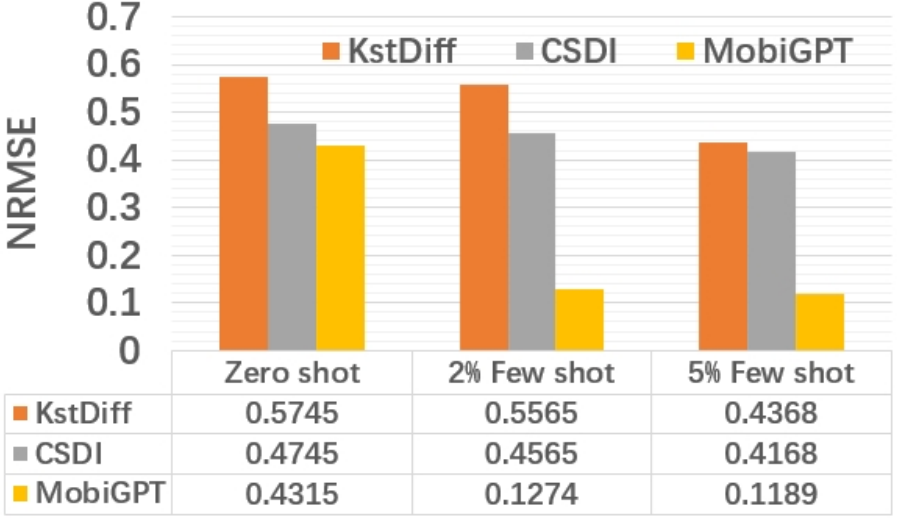}}
    \quad
    \subfloat[Long-term prediction task.]{\includegraphics[width=0.45\linewidth]{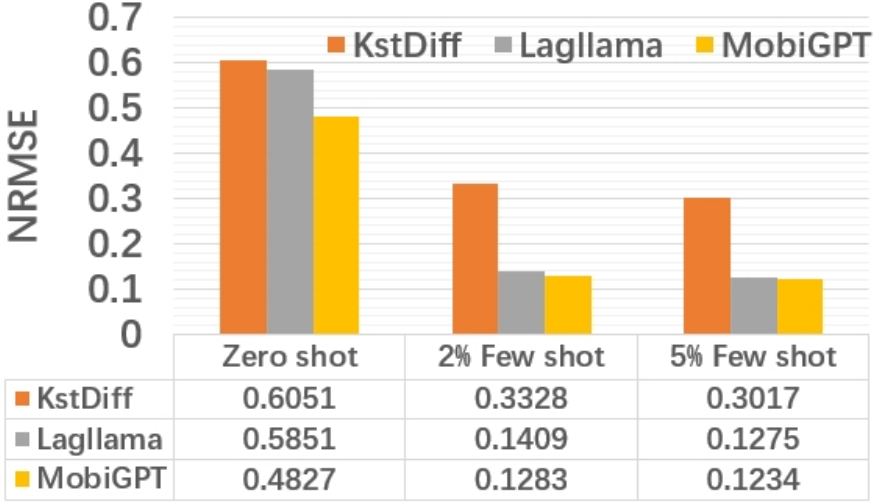}}
    \caption{Zero/Few-shot across two distinct cities.}
    \label{zero_few1}
\end{figure}

Zero/few-shot capabilities are important metrics for evaluating the performance of foundational models. We use BS traffic data of one new city in China (Shandong cell-level traffic) that the model has not encountered during training, and test four models: Lagllama, KstDiff, CSDI, and MobiGPT. In the zero-shot test, we perform data inference directly using the pre-trained model, while in the few-shot test, we select 2\% (2\% few-shot) and 5\% (5\% few-shot) of the new dataset for model fine-tuning, followed by data inference, where the experimental results are as shown in Figure~\ref{zero_few1}.
MobiGPT outperforms other baselines in zero-shot performance for long-term forecasting and generation tasks, with a maximum 21.51\% improvement (from 0.5565 of CSDI to 0.4368), indicating that our model can effectively forecast network traffic in unseen scenarios. Furthermore, after training with a small number of samples, the model's forecasting accuracy improves significantly. This demonstrates that MobiGPT can quickly learn the data characteristics of new scenarios, enhancing its performance.

\begin{table}[t]
\caption{Performance of ablation study. }
\label{ablation}
\renewcommand{\arraystretch}{1.2}
{
\resizebox{0.5\textwidth}{!}{
\begin{tabular}{c|cc|cc|cc}
\hline
\textbf{Model}        & \multicolumn{2}{c|}{\textbf{BS traffic}} & \multicolumn{2}{c|}{\textbf{App traffic}} & \multicolumn{2}{c}{\textbf{Channel}} \\ \hline
\textbf{Metric}             & Generation        & Prediction        & Generation         & Prediction        & Generation        & Prediction        \\
\multicolumn{1}{l|}{} & (JSD)            & (NRMSE)             & (JSD)             & (NRMSE)             & (JSD)            & (NRMSE)             \\ \hline
\textbf{w/o prompt}   & 0.0323           & 0.1809          & 0.2704           & 0.1481           & 0.00147           & 0.1434           \\ \hline
\textbf{w/o period}    & \underline{0.0311}           & 0.1766           & 0.1815            & 0.1472           & 0.00125          & 0.1319           \\ \hline
\textbf{w/o temporal}     & 0.0282           &  \underline{0.1799}         &  \underline{0.2193}           & 0.1415           & 0.00121         & 0.1351          \\ \hline
\textbf{w/o feature}    & 0.0288           & 0.1760         & 0.1913            &  \underline{0.1491}           &  \underline{0.00144}        &  \underline{0.1393}         \\ \hline
\textbf{MobiGPT}    & 0.0281          & 0.1739           & 0.1739          & 0.1397           & 0.00105          & 0.1200          \\ \hline
\textbf{Degradation (\%)}        & -9.64         & -7.46       & -41.95         & -6.32         & -27.08        & -13.86         \\ \hline
\end{tabular}
}
}
\end{table}

\begin{figure}[ht]
  \centering
  \includegraphics[width=0.7\linewidth]{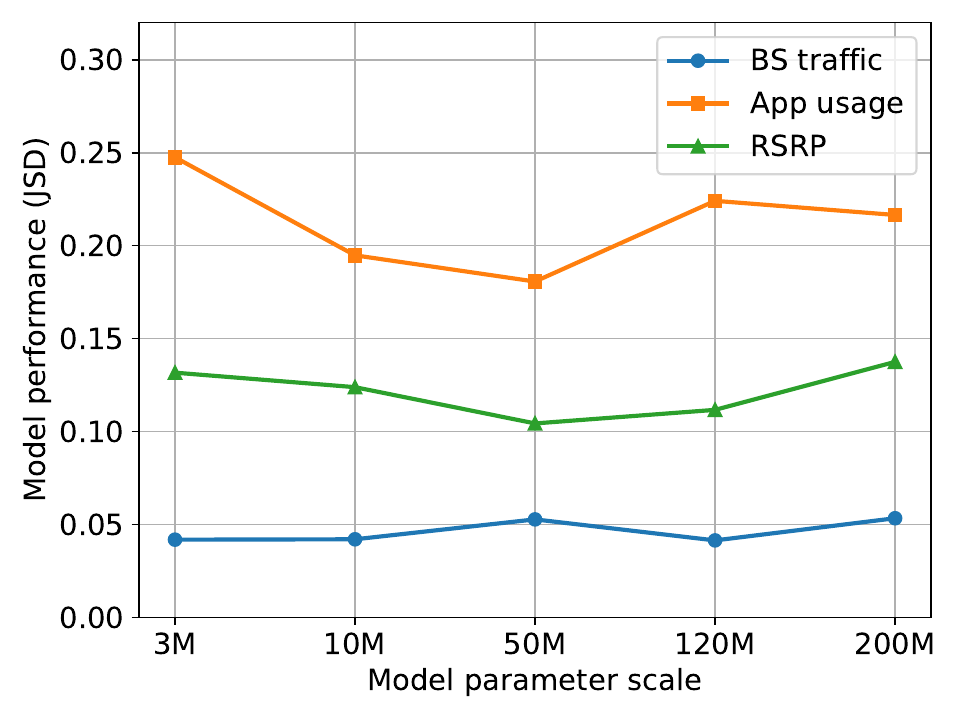}
  \caption{Scaling performance of MobiGPT.}
  \label{radar}
\end{figure}

Next, we test the role of the prompt network. Specifically, we remove different modules from the domain-based prompt network and test MobiGPT's forecasting performance. The results are shown in Table~\ref{ablation}, where 'w/o prompt' indicates the removal of the entire prompt network, and 'w/o period', 'w/o temporal', and 'w/o feature' represent the removal of the periodical, temporal, and feature soft-prompts, respectively. The \underline{text} indicates the module with the most significant impact on the model. It can be seen that all soft prompts in the prompt network contribute differently to improving the model's performance. For BS traffic, which exhibits clear periodicity, the periodical and temporal prompts are the most helpful, and removing these modules results in a decrease of 9.64\% and 7.64\%, respectively. Unlike BS traffic, app traffic does not exhibit obvious periodic patterns, wherein the temporal and feature prompts have the greatest impact. For RSRP, the feature dimension prompt is more critical, as RSRP is closely related to engineering parameters at each time point, which is converted to the feature dimensions $C_0$ in $\mathbb{R}^{B, L, C_0}$.

We further test the model's scaling performance, where we increase the number of transformer blocks in Figure~\ref{foundationmodel} and the dimension of channel $C_0$, crafting five models with different parameter scales (5M, 10M, 50M, 120M, and 200M).
The results are shown in Figure 2. It can be figured that under limited data conditions, model performance does not scale proportionally with the number of parameters. The best performance is achieved around the 50M scale. This preliminary result reveals the scaling law of MobiGPT: under limited data conditions, increasing model parameters is not necessarily the best choice. When model parameters become excessively large, overfitting can occur, which reduces the overall performance of the model. This finding guides future optimization of MobiGPT, helping us make trade-offs and optimization decisions between data volume and parameter scale

\section{Conclusion and future work}

In this paper, we propose MobiGPT as a foundational model for forecasting various types of data and tasks in mobile networks. MobiGPT employs temporal masking techniques to navigate the model through three types of tasks: short-term, long-term, and generation, achieving flexible data forecasting with fixed input data length. We also design a domain-based prompt network, where three types of data-driven soft prompts are used to learn the inherent periodicity and temporal correlations within the data. Furthermore, we introduce a prompting method for extracting environmental semantic features, which effectively learns the dependencies between the mobile environments and different types of mobile data.
We test our MobiGPT's performance using over 100,000 real-world data, and the results show that our method can significantly improve forecasting accuracy for multiple data types by over 20\%. Additionally, MobiGPT demonstrates excellent transferability in unseen scenarios, showcasing its generalization and robustness.

Unlike natural language-based LLMs, MobiGPT is designed from the perspective of time-series data, offering stronger capabilities in understanding and predicting temporal features. This model is well-suited to the mobile communications domain, which predominantly deals with time-series data, and holds potential for adaptation to various scenarios such as sea surface, low-altitude, and underwater scenarios.
In future work, we will continue to incorporate more types of mobile data into MobiGPT, such as grid-level traffic, PRB utilization, user mobility trajectories, service usage sequences, and end-to-end throughput. Moreover, the design of the prompt network is crucial to MobiGPT. While the current model explores only periodicity and temporal correlations, future research will focus on enhancing the specialization and domain adaptation of MobiGPT via integrating spatial attributes and communication prior knowledge to the prompt networks.




\bibliographystyle{gbt7714-numerical}
\bibliography{myref}

\begin{thebibliography}{41}
\providecommand{\natexlab}[1]{#1}
\providecommand{\url}[1]{#1}
\expandafter\ifx\csname urlstyle\endcsname\relax\else
  \urlstyle{same}\fi
\expandafter\ifx\csname href\endcsname\relax
  \DeclareUrlCommand\doi{\urlstyle{rm}}
  \def\eprint#1#2{#2}
\else
  \def\doi#1{\href{https://doi.org/#1}{\nolinkurl{#1}}}
  \let\eprint\href
\fi

\bibitem[Zhang et~al.(2024)Zhang, Han, Xu, Ni, Liu, and Xiong]{zhang2024urban}
ZHANG W, HAN J, XU Z, et~al.
\newblock Urban foundation models: A survey\allowbreak[C]//\allowbreak
Proceedings of the 30th ACM SIGKDD Conference on Knowledge Discovery and Data Mining.
\newblock 2024: 6633-6643.

\bibitem[Zhang et~al.(2024)Zhang, Zhao, Xia, Sun, Sun, Qin, Li, Zhao, Zhao, Cai, et~al.]{zhang2024multimodal}
ZHANG W, ZHAO L, XIA H, et~al.
\newblock A multimodal foundation agent for financial trading: Tool-augmented, diversified, and generalist\allowbreak[C]//\allowbreak
Proceedings of the 30th ACM SIGKDD Conference on Knowledge Discovery and Data Mining.
\newblock 2024: 4314-4325.

\bibitem[Goertzel(2014)]{goertzel2014artificial}
GOERTZEL B.
\newblock Artificial general intelligence: concept, state of the art, and future prospects\allowbreak[J].
\newblock Journal of Artificial General Intelligence, 2014, 5\allowbreak (1): 1.

\bibitem[Ding et~al.(2024)Ding, Zhang, Shang, Zhang, Zong, Feng, Yuan, Su, Li, Sukiennik, Xu, and Li]{ding2024understandingworldpredictingfuture}
DING J, ZHANG Y, SHANG Y, et~al.
\newblock Understanding world or predicting future? a comprehensive survey of world models\allowbreak[A].
\newblock 2024.
\newblock arXiv: \eprint{https://arxiv.org/abs/2411.14499}{2411.14499}.

\bibitem[Pernici et~al.(2006)Pernici and Krogstie]{pernici2006mobile}
PERNICI B, KROGSTIE J.
\newblock Mobile information systems\allowbreak[M].
\newblock Springer, 2006.

\bibitem[Nguyen et~al.(2021)Nguyen, Trestian, To, and Tatipamula]{nguyen2021digital}
NGUYEN H~X, TRESTIAN R, TO D, et~al.
\newblock Digital twin for 5g and beyond\allowbreak[J].
\newblock IEEE Communications Magazine, 2021, 59\allowbreak (2): 10-15.

\bibitem[Yang et~al.(2014)Yang, Huang, Gao, Chang, and Xie]{6760605}
YANG H, HUANG A, GAO R, et~al.
\newblock Interference self-coordination: A proposal to enhance reliability of system-level information in ofdm-based mobile networks via pci planning\allowbreak[J/OL].
\newblock IEEE Transactions on Wireless Communications, 2014, 13\allowbreak (4): 1874-1887.
\newblock DOI: \doi{10.1109/TWC.2014.030514.130447}.

\bibitem[Ji et~al.(2024)Ji, Zhou, Sheng, Li, and Han]{10478861}
JI S, ZHOU D, SHENG M, et~al.
\newblock Dynamic space-ground integrated mobility management strategy for mega leo satellite constellations\allowbreak[J/OL].
\newblock IEEE Transactions on Wireless Communications, 2024, 23\allowbreak (9): 11043-11060.
\newblock DOI: \doi{10.1109/TWC.2024.3378293}.

\bibitem[Xu et~al.(2023)Xu, Kishk, and Alouini]{xu2023space}
XU J, KISHK M~A, ALOUINI M~S.
\newblock Space-air-ground-sea integrated networks: Modeling and coverage analysis\allowbreak[J].
\newblock IEEE Transactions on Wireless Communications, 2023, 22\allowbreak (9): 6298-6313.

\bibitem[Chai et~al.(2025)Chai, Qi, and Li]{10836819}
CHAI H, QI X, LI Y.
\newblock Spatio-temporal knowledge driven diffusion model for mobile traffic generation\allowbreak[J/OL].
\newblock IEEE Transactions on Mobile Computing, 2025: 1-18.
\newblock DOI: \doi{10.1109/TMC.2025.3527966}.

\bibitem[Kanto et~al.(2024)Kanto and Watabe]{10575638}
KANTO Y, WATABE K.
\newblock Wireless link quality estimation using lstm model\allowbreak[C/OL]//\allowbreak
NOMS 2024-2024 IEEE Network Operations and Management Symposium.
\newblock 2024: 1-5.
\newblock DOI: \doi{10.1109/NOMS59830.2024.10575638}.

\bibitem[Khaokaew et~al.(2021)Khaokaew, Rahaman, White, and Salim]{10.1145/3459637.3482076}
KHAOKAEW Y, RAHAMAN M~S, WHITE R~W, et~al.
\newblock Cosem: Contextual and semantic embedding for app usage prediction\allowbreak[C/OL]//\allowbreak
CIKM '21: Proceedings of the 30th ACM International Conference on Information \& Knowledge Management.
\newblock New York, NY, USA: Association for Computing Machinery, 2021: 3137–3141.
\newblock DOI: \doi{10.1145/3459637.3482076}.

\bibitem[Cai et~al.(2023)Cai, Wang, Hui, Chen, Yue, Wang, Zhang, Cheng, and Li]{9896145}
CAI X, WANG L, HUI Y, et~al.
\newblock Coverage optimization for directional sensor networks: A novel sensor redeployment scheme\allowbreak[J/OL].
\newblock IEEE Internet of Things Journal, 2023, 10\allowbreak (2): 1461-1475.
\newblock DOI: \doi{10.1109/JIOT.2022.3208056}.

\bibitem[Han et~al.(2024)Han, Yu, Bai, Wang, Choi, and Zhang]{10242332}
HAN R, YU Y, BAI L, et~al.
\newblock Effective capacity analysis of delay-sensitive communications in noma systems\allowbreak[J/OL].
\newblock IEEE Transactions on Wireless Communications, 2024, 23\allowbreak (4): 3665-3675.
\newblock DOI: \doi{10.1109/TWC.2023.3309958}.

\bibitem[Meskar et~al.(2023)Meskar and Liang]{9923420}
MESKAR E, LIANG B.
\newblock Fair multi-resource allocation in heterogeneous servers with an external resource type\allowbreak[J/OL].
\newblock IEEE/ACM Transactions on Networking, 2023, 31\allowbreak (3): 1244-1262.
\newblock DOI: \doi{10.1109/TNET.2022.3213426}.

\bibitem[Zhu et~al.(2022)Zhu, Li, Zheng, Xia, Xu, Li, Di, Zhi, and Cheng]{10063388}
ZHU X, LI Y, ZHENG Y, et~al.
\newblock Research on 5g network capacity and expansion\allowbreak[C/OL]//\allowbreak
2022 IEEE International Conference on Trust, Security and Privacy in Computing and Communications (TrustCom).
\newblock 2022: 1473-1478.
\newblock DOI: \doi{10.1109/TrustCom56396.2022.00209}.

\bibitem[Wu et~al.(2023)Wu, Jin, Zhang, Cai, and Ji]{10390348}
WU Y, JIN P, ZHANG Y, et~al.
\newblock Coverage quality optimization strategy for static heterogeneous wireless sensor networks\allowbreak[C/OL]//\allowbreak
2023 International Conference on Ambient Intelligence, Knowledge Informatics and Industrial Electronics (AIKIIE).
\newblock 2023: 1-6.
\newblock DOI: \doi{10.1109/AIKIIE60097.2023.10390348}.

\bibitem[Mei et~al.(2023)Mei and Zhang]{10086045}
MEI W, ZHANG R.
\newblock Joint base station and irs deployment for enhancing network coverage: A graph-based modeling and optimization approach\allowbreak[J/OL].
\newblock IEEE Transactions on Wireless Communications, 2023, 22\allowbreak (11): 8200-8213.
\newblock DOI: \doi{10.1109/TWC.2023.3260805}.

\bibitem[Bonald(2006)]{bonald2006erlang}
BONALD T.
\newblock The erlang model with non-poisson call arrivals\allowbreak[J].
\newblock ACM SIGMETRICS Performance Evaluation Review, 2006, 34\allowbreak (1): 276-286.

\bibitem[Li et~al.(2023)Li, Jin, Li, Huang, Ma, Cui, Huang, Qiao, and Yoo]{10.1145/3586164}
LI H, JIN D, LI X, et~al.
\newblock Dmgf-net: An efficient dynamic multi-graph fusion network for traffic prediction\allowbreak[J/OL].
\newblock ACM Trans. Knowl. Discov. Data, 2023, 17\allowbreak (7).
\newblock DOI: \doi{10.1145/3586164}.

\bibitem[Zhou et~al.(2023)Zhou, Ding, Liu, Jin, and Li]{10.1145/3589132.3625641}
ZHOU Z, DING J, LIU Y, et~al.
\newblock Towards generative modeling of urban flow through knowledge-enhanced denoising diffusion\allowbreak[C/OL]//\allowbreak
SIGSPATIAL '23: Proceedings of the 31st ACM International Conference on Advances in Geographic Information Systems.
\newblock New York, NY, USA: Association for Computing Machinery, 2023.
\newblock DOI: \doi{10.1145/3589132.3625641}.

\bibitem[Lin et~al.(2020)Lin, Jain, Wang, Fanti, and Sekar]{lin2020using}
LIN Z, JAIN A, WANG C, et~al.
\newblock Using gans for sharing networked time series data: Challenges, initial promise, and open questions\allowbreak[C]//\allowbreak
Proceedings of the ACM Internet Measurement Conference.
\newblock 2020: 464-483.

\bibitem[Yin et~al.(2022)Yin, Lin, Jin, Fanti, and Sekar]{yin2022practical}
YIN Y, LIN Z, JIN M, et~al.
\newblock Practical gan-based synthetic ip header trace generation using netshare\allowbreak[C]//\allowbreak
Proceedings of the ACM SIGCOMM 2022 Conference.
\newblock 2022: 458-472.

\bibitem[Wang et~al.(2020)Wang, Li, Ye, Wang, and Zhang]{wang2020packetcgan}
WANG P, LI S, YE F, et~al.
\newblock Packetcgan: Exploratory study of class imbalance for encrypted traffic classification using cgan\allowbreak[C]//\allowbreak
ICC 2020-2020 IEEE International Conference on Communications (ICC).
\newblock IEEE, 2020: 1-7.

\bibitem[Wu et~al.(2022)Wu, Ma, Ye, Zhang, Shao, and Zheng]{9700950}
WU S, MA B, YE T, et~al.
\newblock A machine learning based intelligent propagation model for rsrp prediction\allowbreak[C/OL]//\allowbreak
2022 International Seminar on Computer Science and Engineering Technology (SCSET).
\newblock 2022: 1-5.
\newblock DOI: \doi{10.1109/SCSET55041.2022.00010}.

\bibitem[Tao et~al.(2020)Tao, Lu, liang Wang, and Qian]{Tao2020FeatureEB}
TAO L, LU J, LIANG WANG X, et~al.
\newblock Feature engineering based intelligent wireless propagation model for rsrp prediction\allowbreak[J].
\newblock IOP Conference Series: Materials Science and Engineering, 2020, 768.

\bibitem[Yang et~al.(2023)Yang, Liu, and Wang]{yang2023fingpt}
YANG H, LIU X~Y, WANG C~D.
\newblock Fingpt: Open-source financial large language models\allowbreak[J].
\newblock FinLLM Symposium at IJCAI 2023, 2023.

\bibitem[Zhang et~al.(2024)Zhang, Han, Xu, Ni, Liu, and Xiong]{Zhang2024TowardsUG}
ZHANG W, HAN J, XU Z, et~al.
\newblock Towards urban general intelligence: A review and outlook of urban foundation models: abs/2402.01749\allowbreak[A].
\newblock 2024.

\bibitem[Cao et~al.(2024)Cao, Jia, Arik, Pfister, Zheng, Ye, and Liu]{cao2024tempopromtrained}
CAO D, JIA F, ARIK S~O, et~al.
\newblock Tempo: Prompt-based generative pre-trained transformer for time series forecasting\allowbreak[A].
\newblock 2024.
\newblock arXiv: \eprint{https://arxiv.org/abs/2310.04948}{2310.04948}.

\bibitem[Jin et~al.(2024)Jin, Wang, Ma, Chu, Zhang, Shi, Chen, Liang, Li, Pan, and Wen]{jin2024timellmtimeseriesforecasting}
JIN M, WANG S, MA L, et~al.
\newblock Time-llm: Time series forecasting by reprogramming large language models\allowbreak[A].
\newblock 2024.
\newblock arXiv: \eprint{https://arxiv.org/abs/2310.01728}{2310.01728}.

\bibitem[Rasul et~al.(2024)Rasul, Ashok, Williams, Ghonia, Bhagwatkar, Khorasani, Bayazi, Adamopoulos, Riachi, Hassen, Biloš, Garg, Schneider, Chapados, Drouin, Zantedeschi, Nevmyvaka, and Rish]{rasul2024lagllamafoundationmodelsprobabilistic}
RASUL K, ASHOK A, WILLIAMS A~R, et~al.
\newblock Lag-llama: Towards foundation models for probabilistic time series forecasting\allowbreak[A].
\newblock 2024.
\newblock arXiv: \eprint{https://arxiv.org/abs/2310.08278}{2310.08278}.

\bibitem[Yuan et~al.(2024)Yuan, Ding, Feng, Jin, and Li]{10.1145/3637528.3671662}
YUAN Y, DING J, FENG J, et~al.
\newblock Unist: A prompt-empowered universal model for urban spatio-temporal prediction\allowbreak[C/OL]//\allowbreak
KDD '24: Proceedings of the 30th ACM SIGKDD Conference on Knowledge Discovery and Data Mining.
\newblock New York, NY, USA: Association for Computing Machinery, 2024: 4095–4106.
\newblock DOI: \doi{10.1145/3637528.3671662}.

\bibitem[Garza et~al.(2023)Garza and Mergenthaler-Canseco]{garza2023timegpt1}
GARZA A, MERGENTHALER-CANSECO M.
\newblock Timegpt-1\allowbreak[A].
\newblock 2023.
\newblock arXiv: \eprint{https://arxiv.org/abs/2310.03589}{2310.03589}.

\bibitem[Liu et~al.(2023)Liu, Ding, Fu, and Li]{liu2023urbankg}
LIU Y, DING J, FU Y, et~al.
\newblock Urbankg: An urban knowledge graph system\allowbreak[J].
\newblock ACM Transactions on Intelligent Systems and Technology, 2023, 14\allowbreak (4): 1-25.

\bibitem[Li et~al.(2023)Li, Li, Zhang, Tarkoma, and Hui]{10.1145/3597212}
LI T, LI Y, ZHANG M, et~al.
\newblock You are how you use apps: User profiling based on spatiotemporal app usage behavior\allowbreak[J/OL].
\newblock ACM Trans. Intell. Syst. Technol., 2023, 14\allowbreak (4).
\newblock DOI: \doi{10.1145/3597212}.

\bibitem[Peebles et~al.(2023)Peebles and Xie]{peebles2023scalable}
PEEBLES W, XIE S.
\newblock Scalable diffusion models with transformers\allowbreak[C]//\allowbreak
Proceedings of the IEEE/CVF International Conference on Computer Vision.
\newblock 2023: 4195-4205.

\bibitem[Zheng(2020)]{4ba2-tg21-20}
ZHENG Y.
\newblock Rsrpset: the dataset of the 16th cpgmcm\allowbreak[M/OL].
\newblock IEEE Dataport, 2020.
\newblock DOI: \doi{10.21227/4ba2-tg21}.

\bibitem[Gong et~al.(2023)Gong, Liu, Li, Chai, Wang, Feng, Deng, Jin, and Li]{10.1145/3589132.3625569}
GONG J, LIU Y, LI T, et~al.
\newblock Empowering spatial knowledge graph for mobile traffic prediction\allowbreak[C/OL]//\allowbreak
SIGSPATIAL '23: Proceedings of the 31st ACM International Conference on Advances in Geographic Information Systems.
\newblock New York, NY, USA: Association for Computing Machinery, 2023.
\newblock DOI: \doi{10.1145/3589132.3625569}.

\bibitem[Pandey et~al.(2024)Pandey, Tiwari, Rodrigues, and Roy]{10449458}
PANDEY C, TIWARI V, RODRIGUES J~J~P~C, et~al.
\newblock 5gt-gan-net: Internet traffic data forecasting with supervised loss based synthetic data over 5g\allowbreak[J/OL].
\newblock IEEE Transactions on Mobile Computing, 2024, 23\allowbreak (11): 10694-10705.
\newblock DOI: \doi{10.1109/TMC.2024.3364655}.

\bibitem[Madane et~al.(2022)Madane, djallel Dilmi, Forest, Azzag, Lebbah, and Lacaille]{madane2022transformerbasedconditionalgenerativeadversarial}
MADANE A, DJALLEL DILMI M, FOREST F, et~al.
\newblock Transformer-based conditional generative adversarial network for multivariate time series generation\allowbreak[A].
\newblock 2022.
\newblock arXiv: \eprint{https://arxiv.org/abs/2210.02089}{2210.02089}.

\bibitem[Tashiro et~al.(2021)Tashiro, Song, Song, and Ermon]{10.5555/3540261.3542161}
TASHIRO Y, SONG J, SONG Y, et~al.
\newblock Csdi: conditional score-based diffusion models for probabilistic time series imputation\allowbreak[C]//\allowbreak
NIPS '21: Proceedings of the 35th International Conference on Neural Information Processing Systems.
\newblock Red Hook, NY, USA: Curran Associates Inc., 2021.

\end{thebibliography}

\clearpage
\biographies

\begin{CCJNLbiography}{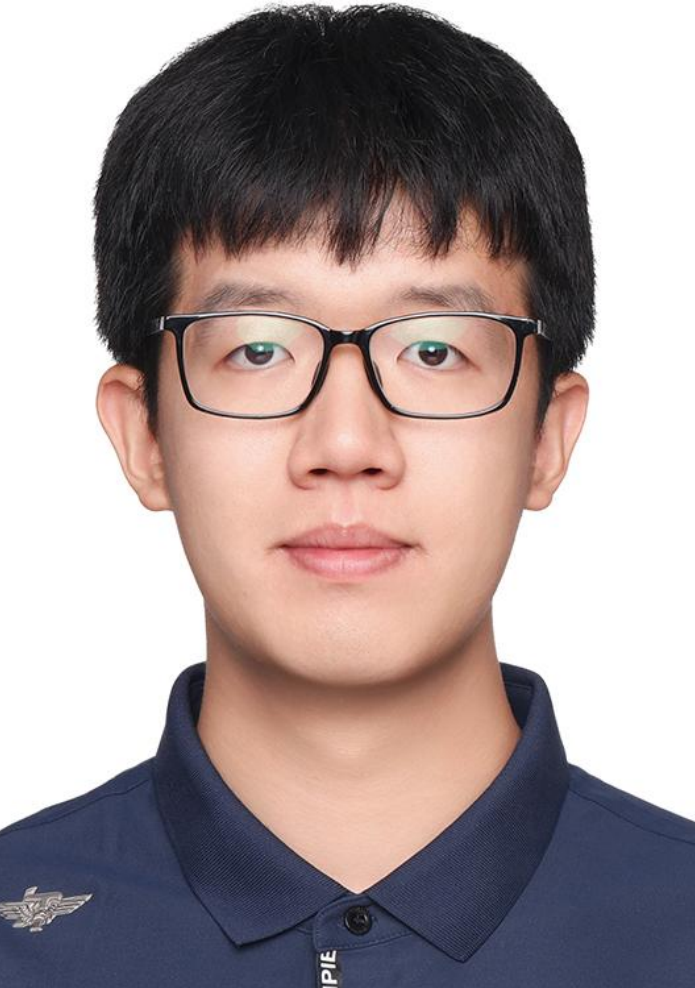}{Xiaoqian Qi}
received the B.S. degree in Electronic and Information Engineering from the School of Information and Electronics, Beijing Institute of Technology in 2024. At present, he is studying for the M.S. degree in the Department of Electronic Engineering, Tsinghua
University. His research interests include spatio-temporal data prediction and generation, mobile big data mining, AI for communication, and urban computing.
\end{CCJNLbiography}

\begin{CCJNLbiography}{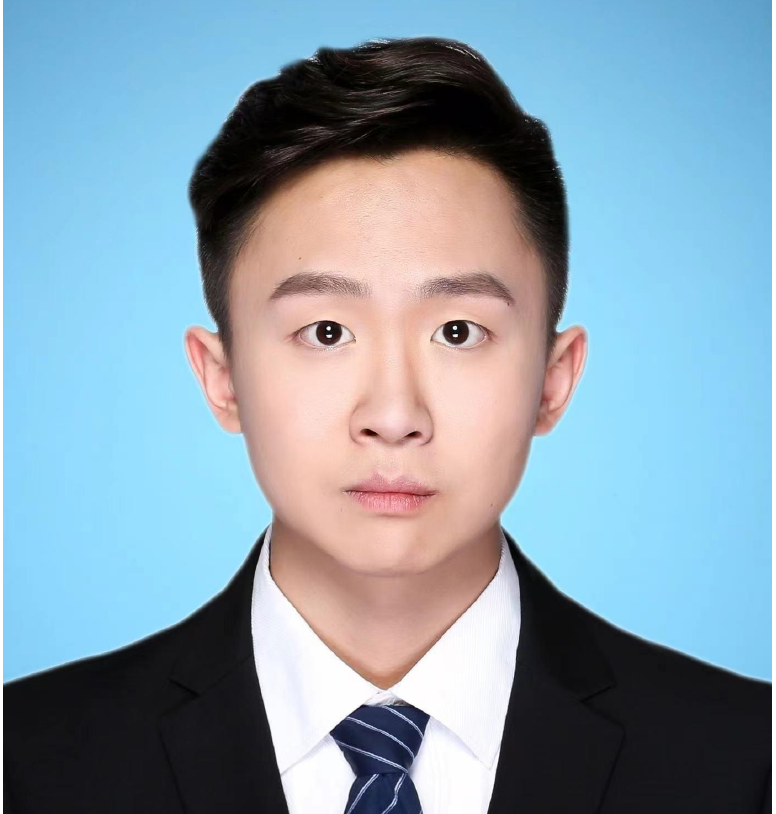}{Haoye Chai}
is currently an assistant researcher of the Department of Electronic Engineering, Tsinghua University. He received his Ph.D. and B.S. degrees from the University of Electronic Science and Technology of China in 2022 and 2016, respectively. His research interests include mobile networks, generative AI, and spatio-temporal data mining.

\end{CCJNLbiography}

\begin{CCJNLbiography}{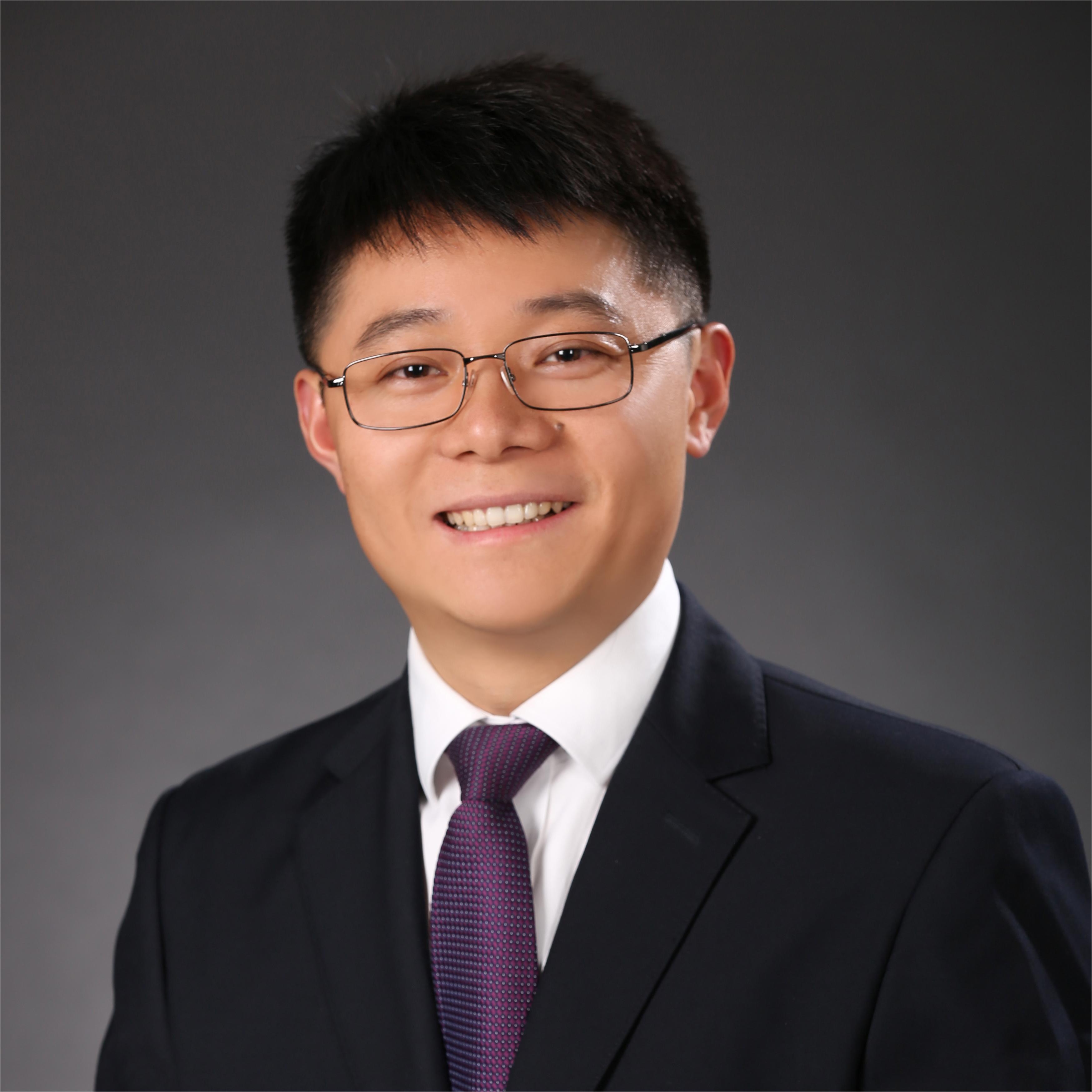}{Yong Li}
is currently a Tenured Professor of the Department of
Electronic Engineering, Tsinghua University. He received the Ph.D. degree in electronic
engineering from Tsinghua University in 2012. His research interests lie in the cutting-edge
interdisciplinary field of artificial intelligence and urban science. Prof. Li has published over
100 papers in first-tier international conferences and journals, including Nature
Computational Science, Nature Machine Intelligence, Nature Cities, Nature Human Behaviour,
KDD, WWW, TMC, UbiComp, AAAI, etc., and his papers have total citations of more than 26,000.
He has received 7 Best/Outstanding Paper Awards from major international conferences in
computer science and has served over 30 times as an SPC/PC member for international conferences such as KDD,
WWW, AAAI, and IJCAI and as a (guest) editorial board member for journals including ACM
IMWUT, IEEE JSAC, and TNSM. He has been recognized as a "Highly Cited Researcher" and
selected for the national "Ten Thousand Talents Program" as a Youth Top-Notch Talent.

\end{CCJNLbiography}

\end{document}